%% file: emnlp2026.tex
\pdfoutput=1

\documentclass[11pt]{article}

\usepackage{EMNLP2026}

\usepackage{times}
\usepackage{latexsym}

\usepackage[T1]{fontenc}

\usepackage[utf8]{inputenc}
\usepackage{amsmath}
\usepackage{microtype}

\usepackage{inconsolata}
\usepackage{xcolor}
\usepackage{graphicx}
\usepackage{subcaption}
\usepackage{listings}
\usepackage{booktabs}
\usepackage{multirow}

\newcommand{\ignore}[1]{}
\definecolor{lightyellow}{RGB}{255, 255, 150}
\definecolor{lightblue}{RGB}{179, 230, 255}
\definecolor{lightgray}{RGB}{220, 220, 220}

\title{Post-Training Large Language Models with LLMON}

\author{
  Basel Shbita\textsuperscript{1},
  Michael Hind\textsuperscript{2},
  Chad DeLuca\textsuperscript{1},
  Nathan Fulton\textsuperscript{3},
  Shubhi Asthana\textsuperscript{1},\\
  \textbf{Farhan Ahmed\textsuperscript{1},
  Anisa Halimi\textsuperscript{4},
  Yada Zhu\textsuperscript{2},
  Mauro Martino\textsuperscript{3},
  David Cox\textsuperscript{3},
  Dan Gutfreund\textsuperscript{3}}\\
  \textsuperscript{1}IBM Research, San Jose, CA, USA \\
  \textsuperscript{2}IBM Research, Yorktown Heights, NY, USA \\
  \textsuperscript{3}IBM Research, Cambridge, MA, USA \\
  \textsuperscript{4}IBM Research, Dublin, Ireland \\
  \small{\texttt{basel@ibm.com},
  \texttt{\{hindm, delucac\}@us.ibm.com},
  \texttt{nathan@ibm.com},
  \texttt{sasthan@us.ibm.com},}\\
  \small{\texttt{\{farhan.ahmed, anisa.halimi\}@ibm.com},
  \texttt{\{yzhu, mmartino\}@us.ibm.com},
  \texttt{david.d.cox@ibm.com},
  \texttt{dgutfre@us.ibm.com}}
}

\begin{document}
\maketitle

\input{tex/00_abstract}
\input{tex/01_introduction}
\input{tex/02_related_work}
\input{tex/03_llmon}
\input{tex/04_exps}
\input{tex/05_results}
\input{tex/06_conclusion}
\input{tex/07_limitations}

\bibliography{emnlp2026}
\bibliographystyle{acl_natbib}

\appendix
\input{tex/08_appendix}

\end{document}

%% file: tex/00_abstract.tex
\begin{abstract}
Large Language Models (LLMs) treat all context tokens uniformly, lacking explicit separation between executable instructions and passive data.
This leads to prompt injection vulnerabilities, ambiguous instruction binding, and unreliable execution control.
We present a post-training methodology using LLMON (LLM Object Notation), a framework that enables separating instructions from data and requires explicit identifier-based execution.
We construct a large-scale structured training corpus of 4.6B tokens spanning general, RAG (Retrieval-Augmented Generation), and tool-calling data, augmented with distractor-infused variants in which competing instructions and irrelevant data segments are present in context but not referenced.
Post-training two Granite-4.1 base models (3B and 8B) through supervised fine-tuning followed by reinforcement learning, we show that LLMON-structured supervision reduces attack success rates on indirect prompt-injection benchmarks from 53-100\% to 0.02-5.6\% while preserving general capabilities.
We further introduce attention-based interpretability metrics that diagnose failure modes at individual transformer layers for structured inputs, revealing how our approach reshapes internal attention patterns.
Our results demonstrate that explicit execution semantics can be taught through standard post-training without architectural modification.
\end{abstract}


%% file: tex/01_introduction.tex
\section{Introduction}
\label{sec:intro}

Large Language Models (LLMs) process all tokens in a context window through the same attention mechanism, without any built-in distinction between executable instructions, passive data, and auxiliary context~\citep{vaswani2017attention, kaddour2023challenges}.
This architectural uniformity means that any token can equally influence model behavior, whether it originates from a user instruction, a retrieved document, or an injected adversarial payload.
The practical consequences are well-documented: prompt injection attacks that hijack model execution~\citep{greshake2023promptinjection, perez2022ignore}, ambiguous referential binding when multiple instructions coexist in context (e.g., ``execute the second task''), and degraded reliability in compositional scenarios involving retrieval-augmented generation (RAG) and tool calling~\citep{yao2022react, schick2023toolformer}.
These failure modes stem from a single root cause: the model has no mechanism to distinguish executable instructions from passive data at the token level.
Current instruction-tuning approaches~\citep{ouyang2022instructgpt, wei2022zeroshot} train models to follow instructions implicitly, based on position, formatting conventions (e.g., ``chat templates''), and statistical patterns, thus lack the mechanism to distinguish which instruction to execute when multiple candidates exist in a given prompt.

In this paper, we present a structured post-training methodology built on LLMON~\citep{hind2026llmon,llmon-git} (LLM Object Notation), a hierarchical tagging framework that introduces explicit instruction-data separation and identifier-based execution binding into the LLM interface.
Under LLMON, each instruction and data segment is wrapped in typed, uniquely identified blocks, and execution occurs only through explicit execution calls that reference specific instruction and data identifiers.
This design steers the model toward a clear semantic principle where only explicitly invoked instructions are executed.
Our contributions are:
\begin{enumerate}
\item We describe a scalable data pipeline that transforms instruction-tuning corpora (general, RAG, and tool-calling) into LLMON-structured training data, including distractor-augmented variants with up to three competing instructions and irrelevant data segments.
\item We conduct a post-training evaluation on two base models (\textit{Granite-4.1-3B-Base} and \textit{Granite-4.1-8B-Base}) under full fine-tuning with 4.6B tokens, followed by reinforcement learning.
\item We introduce attention-based interpretability metrics: Instruction Focus Ratio (IFR), Data Attention Score (DAS), and Leakage (LEAK), to provide mechanistic insight into how structured training reshapes model attention.
\end{enumerate}

%% file: tex/02_related_work.tex
\section{Related Work}
\label{sec:related_work}

\paragraph{Instruction tuning and alignment.}
Instruction tuning~\citep{wei2022zeroshot, ouyang2022instructgpt} and preference-based alignment~\citep{rafailov2023dpo, christiano2017preferences} have become the standard post-training pipeline for modern LLMs.
These methods train models to follow natural-language instructions but rely on implicit positional cues and chat templates~\citep{hf_chat_templating}-based formatting conventions, rather than explicit structural semantics.
When multiple instructions appear in context, models lack a principled mechanism for disambiguation.
Our work is complementary: LLMON-structured supervision can be applied on top of any instruction-tuning corpus to add explicit execution binding behavior.

\paragraph{Prompt injection and adversarial robustness.}
Prompt injection exploits the absence of privilege separation between instructions and data~\citep{greshake2023promptinjection, perez2022ignore}.
Spotlighting~\citep{hines2024defending} marks untrusted spans with in-band text delimiters at the prompt level, while StruQ~\citep{chen2025struq} and SecAlign~\citep{chen2025secalign} make the boundary a \emph{training} objective, fine-tuning on structured prompts and preference pairs in which injected instructions are present but must not be followed.
We differ in two respects.
First, the boundary is carried by reserved single tokens from a disjoint region of the vocabulary rather than by in-band delimiters or role names, so it cannot be formed by subword composition of ordinary content.
Second, privilege attaches to an explicit binding rather than to a role: every span is individually addressable and authority follows from which pointer references it, so several same-role instructions can coexist with only one bound for execution.

\paragraph{Tool use and function calling.}
Modern LLM systems integrate external tools through structured generation~\citep{schick2023toolformer, patil2024gorilla, yao2022react}.
These approaches define tool schemas and train models to emit structured calls, but the instruction-data boundary within the prompt remains implicit.
LLMON's execution semantics are orthogonal and compatible with tool-calling pipelines: tool definitions, retrieved documents, and user instructions each occupy typed segments with distinct identifiers.

\paragraph{Structured generation and constrained decoding.}
Constrained decoding approaches~\citep{willard2023efficient, zheng2024sglang, LMQL-2023} enforce output structure at inference time through grammar-based token masking.
Prompting frameworks~\citep{dspy-2023, prompt-frameworks-survey-2026, poml-github} and programming-language abstractions~\citep{MTP-2025} orchestrate LLM calls, and markup-based prompt languages impose document structure on the prompt, but none of these modifies the model's internal representation of instruction-data boundaries.
Unlike these inference-time or application-layer approaches, LLMON operates at the \emph{training data} level, teaching models to internalize structural distinctions through supervised learning.

%% file: tex/03_llmon.tex
\section{Using LLMON (LLM Object Notation)}
\label{sec:llmon_framework}

This section describes how we leverage LLMON~\citep{hind2026llmon} to construct structured post-training data that teaches models explicit execution control.
LLMON is a hierarchical tagging framework that introduces instruction-data separation and identifier-based execution binding at the token level, operating within standard autoregressive language modeling without architectural modification.
We briefly summarize the notation relevant to our training methodology, then describe the data construction pipeline.


\subsection{Notation Overview}

LLMON defines six special delimiter tokens (``\texttt{<|open|>}'', ``\texttt{<|close|>}'', ``\texttt{<|open\_end|>}'', ``\texttt{<|self\_close|>}'', ``\texttt{<|:|>}'', ``\texttt{<|.|>}'') that form a typed tagging system.
The key structural elements relevant to training are:

\paragraph{Typed segments.}
All content is wrapped in identified and tagged blocks.
\emph{Instructions} carry role qualifiers (``\texttt{instr<|:|>ID}''), \emph{data} segments hold passive content (``\texttt{data<|:|>ID}''), \emph{requirements} hold output constraints (``\texttt{req<|:|>ID}''), and \emph{documents} use nested paths (``\texttt{doc<|.|>title}'', ``\texttt{doc<|.|>text}'').
\emph{Tool definitions} are flattened so that every leaf of the JSON schema becomes its own addressable sub-field (``\texttt{tool<|.|>name}'', ``\texttt{tool<|.|>parameters<|.|>}\textit{arg}\texttt{<|.|>type}''), and a tool result returns as a ``\texttt{data}'' block like any other input.
A \emph{system} block (``\texttt{system<|:|>ID}'') carries authoritative policy and persists across turns.

\paragraph{Execution binding.}
An execution block explicitly references which instruction to execute, which data to use as input, and where the output should be placed via identifier references as shown in Figure~\ref{lst:llmon_exec_block}.
This design enforces a clear semantic paradigm where multiple instructions can coexist in context without interference, as execution requires referencing a specific identifier regardless of position.

\paragraph{The privilege invariant.}
Execution binding yields the invariant our training targets: content reached through ``\texttt{exec<|.|>instruction}'' or ``\texttt{exec<|.|>system}'' is \emph{authoritative}, whereas everything reached through ``\texttt{exec<|.|>input}'' (data, documents, tool schemas, and tool results) is material to be read, not a command to be followed, however it is phrased.
Blocks that no pointer references are rendered into the context but bound to nothing.
Privilege is therefore a property of the binding rather than of a block's position or role name, and the delimiters that express it are reserved tokens occupying a disjoint region of the vocabulary, so the structure cannot be formed by subword composition of ordinary content.

\newsavebox{\llmonbox}
\begin{figure}[t]
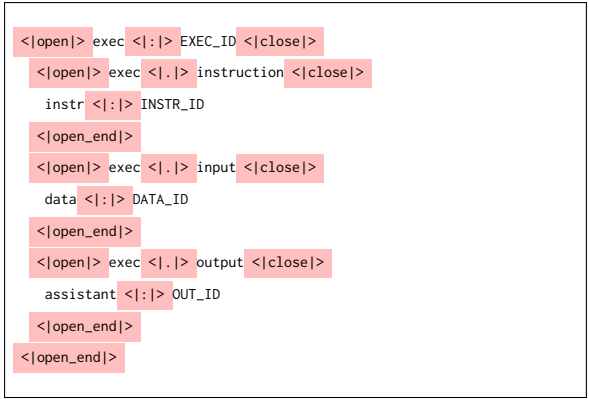

\centering
\begin{lrbox}{\llmonbox}
\begin{minipage}{0.97\linewidth}
\begin{lstlisting}[
    escapechar=@,
    basicstyle=\ttfamily\tiny,
    numbers=none,
    breaklines=true,
    breakindent=0pt,
    showstringspaces=false
]
@\colorbox{pink}{<|open|>}@exec@\colorbox{pink}{<|:|>}@EXEC_ID@\colorbox{pink}{<|close|>}@
  @\colorbox{pink}{<|open|>}@exec@\colorbox{pink}{<|.|>}@instruction@\colorbox{pink}{<|close|>}@
    instr@\colorbox{pink}{<|:|>}@INSTR_ID
  @\colorbox{pink}{<|open\_end|>}@
  @\colorbox{pink}{<|open|>}@exec@\colorbox{pink}{<|.|>}@input@\colorbox{pink}{<|close|>}@
    data@\colorbox{pink}{<|:|>}@DATA_ID
  @\colorbox{pink}{<|open\_end|>}@
  @\colorbox{pink}{<|open|>}@exec@\colorbox{pink}{<|.|>}@output@\colorbox{pink}{<|close|>}@
    assistant@\colorbox{pink}{<|:|>}@OUT_ID
  @\colorbox{pink}{<|open\_end|>}@
@\colorbox{pink}{<|open\_end|>}@
\end{lstlisting}
\end{minipage}
\end{lrbox}
\fbox{\usebox{\llmonbox}}
\caption{Example of an execution block in LLMON syntax. All special tokens are highlighted in light-pink.}
\label{lst:llmon_exec_block}
\end{figure}

\subsection{Data Construction Pipeline}
\label{sec:data_pipeline}

We develop an automated pipeline that transforms diverse instruction-tuning data into LLMON-structured training examples.
The pipeline handles three distinct data modalities:

\paragraph{General instruction data.}
Single-turn and multi-turn conversations are serialized with per-turn instruction blocks, ``\texttt{exec}'' bindings, and assistant data blocks.
Each turn receives unique identifiers, and multi-turn examples chain consecutive ``\texttt{exec}'' cycles.

\paragraph{RAG and document-grounded data.}
Retrieved documents are serialized as typed ``\texttt{doc}'' segments with nested ``\texttt{doc<|.|>title}'' and ``\texttt{doc<|.|>text}'' sub-fields.
The ``\texttt{exec}'' block's input field references both the user's data and the relevant document identifiers, teaching the model to ground its response in explicitly referenced context.

\paragraph{Tool-calling data.}
Tool definitions are serialized using flattened path-based notation (e.g., ``\texttt{tool<|.|>function<|.|>name}'', ``\texttt{tool<|.|>function<|.|>description}'').
Tool invocations produce a first ``\texttt{exec}'' cycle (user instruction $\rightarrow$ assistant tool call), followed by a tool-result data block and a second ``\texttt{exec}'' cycle (tool result $\rightarrow$ final assistant response).


\subsection{Distractor Augmentation}
\label{sec:distractor_aug}

The central innovation in our training data is \emph{distractor augmentation}: systematically injecting competing instructions and irrelevant data into the context while keeping the ``\texttt{exec}'' block pointed at only the target.
This directly trains the model to resolve referential bindings under interference.

For each base example we generate variants with $d \in \{1,2,3\}$ distractor instructions (sampled from the instruction pool, excluding the target) and $m \in \{0,1,2,3\}$ irrelevant data segments (sampled from other examples).
The distractor instructions and data occupy their own identified blocks in the context, but the ``\texttt{exec}'' block references \emph{only} the original target instruction and data, and the target output is identical across all variants of an example.
Variants are sampled per base example, yielding 8 categories parameterized by whether target data is present and by the distractor and irrelevant-data counts.
A model that keys on the ``\texttt{exec}'' block is unaffected by the decoys; one that keys on surface cues such as position or imperative phrasing is penalized by the loss.

We illustrate one such example with a multi-instruction and multi-data distractors execution in Appendix~\ref{sec:appndx_distractor}.





%% file: tex/04_exps.tex
\section{Experiments}
\label{sec:exps}


\subsection{Training Data}
\label{sec:data}

Following the pipeline of Section~\ref{sec:data_pipeline}, we assemble the LLMON-structured corpus from the sources below, totaling 4.4M examples and 4.6B tokens.

\begin{itemize}
\item \textbf{General SFT data}: 3.25M single-turn (2.79B tokens) and 59k multi-turn (120M tokens) examples drawn from the Granite-4.1 post-training SFT corpora.
\item \textbf{RAG data}: 27k single-turn (45.5M tokens) and 8k multi-turn (17.8M tokens) retrieval-augmented examples from the same corpora.
\item \textbf{Tool-calling data}: 451k single-turn (707M tokens) and 251k multi-turn (562M tokens) function-calling examples from the same corpora.
\item \textbf{Alpaca}: 52k clean examples~\citep{taori2023stanford} (17.2M tokens) plus 103k distractor-augmented variants (39.4M tokens).
\item \textbf{Dolly}: 15k clean examples~\citep{conover2023free} (3.3M tokens) plus 30k distractor-augmented variants (16.3M tokens).
\item \textbf{Prompt-injection data}: 130k examples (211M tokens) in which an adversarial instruction is planted inside a payload the model is meant to read rather than execute, spanning general (40k), RAG (11k), and tool-calling (80k) modalities.
\end{itemize}

The general, RAG, and tool-calling examples teach the structural format across single-turn instruction following, multi-turn conversation, document-grounded QA, and tool invocation, and account for 93.6\% of all tokens, which is what keeps capability close to the conventionally-tuned baseline.
The Alpaca and Dolly expansions supply the distractor-augmented variants of Section~\ref{sec:distractor_aug}, expanding 67k base examples into 133k variants.
The prompt-injection examples are the training-time counterpart of the evaluation in Section~\ref{sec:eval_injection}: the injected instruction is rendered as a block that no execution pointer references, and the target output is the one that ignores it.

For the reinforcement-learning stages we use two further sources: approximately 169k prompts across five sequential policy-optimization stages spanning multi-domain reasoning, RLHF, knowledge calibration, and two mathematics boosters; and approximately 1k on-policy preference pairs for the final preference-optimization stage, constructed from each model's own responses at agentic decision points where a tool result carries an injected instruction.


\subsection{Models and Training}
\label{sec:models}

We post-train two base models from the same family and generation at different scales: \textit{Granite-4.1-3B-Base} and \textit{Granite-4.1-8B-Base}~\citep{granite2026granite41,granite2024granite}.
Both are base (pre-trained) checkpoints without prior instruction tuning, so the LLMON behavior is learned during post-training rather than layered on top of an existing chat format.

\paragraph{Supervised fine-tuning.}
We first install the six LLMON delimiters in the tokenizer as reserved single tokens by remapping unused vocabulary slots, which avoids growing the vocabulary or resizing the embedding matrix.
We then run full fine-tuning, updating all parameters on the LLMON-serialized corpus: three epochs over 4.4M examples and 4.6B tokens at a sequence length of 4096 tokens, sharded across 8$\times$H100 GPUs with FSDP.
We report the final-epoch checkpoint.

\paragraph{Reinforcement learning.}
Starting from the supervised checkpoint, we run five sequential Group Relative Policy Optimization (GRPO) stages against a reward model over the $\approx$169k prompts described above, which optimize end behavior directly rather than the token-level target.
A final Direct Preference Optimization~\citep{rafailov2023dpo} (DPO) stage then trains on the $\approx$1k on-policy injection preference pairs ($\beta=0.1$, one epoch, learning rate $5\times10^{-7}$, effective batch size 32).
This stage is placed last so that the general-capability stages, which carry no injection objective, cannot erode the boundary behavior installed earlier.
Unless noted otherwise, results reported as ``LLMON'' refer to this final SFT$\rightarrow$GRPO$\rightarrow$DPO checkpoint.


\subsection{Evaluation}
\label{sec:eval}

\subsubsection{Standard Benchmarks}
\label{sec:eval_standard}

To verify that LLMON post-training preserves general capabilities, we evaluate across seven categories, reporting the unweighted mean over the benchmarks in each:
\textbf{General} (MMLU~\citep{hendrycks2021measuringmathematicalproblemsolving}, MMLU-Pro~\cite{wang2024mmlu}, BBH~\cite{suzgun2023challenging}, AGI-Eval~\cite{zhong2024agieval}, GPQA~\cite{rein2023gpqa});
\textbf{Alignment} (IFEval~\citep{zhou2023instructionfollowingevaluationlargelanguage});
\textbf{Math} (GSM8K~\citep{cobbe2021trainingverifierssolvemath}, GSM-Symbolic~\cite{mirzadeh2025gsm}, Minerva Math~\cite{lewkowycz2022solving}, DeepMind Math~\cite{saxton2019analysing});
\textbf{Code} (HumanEval~\cite{chen2021evaluating}, HumanEval+~\cite{liu2023your}, MBPP~\cite{austin2021program}, MBPP+~\cite{liu2024evaluating}, MultiPL-E~\cite{cassano2023multipl}, CRUXEval-O~\cite{gu2024cruxeval});
\textbf{Tool-Calling} (BFCL v3~\cite{patil2025berkeley});
\textbf{Multilingual} (Global-MMLU~\cite{singh2025global}, MGSM~\cite{shi2022language}, INCLUDE~\cite{romanou2025include});
and \textbf{Safety} (6-task safety aggregate metric from Tulu3~\cite{zhao2026safety}, AttaQ~\cite{kour2023unveiling}, SALAD-Bench~\cite{li2024salad}).
Each LLMON model is compared against the instruct counterpart at the same scale, trained on the matched chat-template corpus of Section~\ref{sec:data}.


\subsubsection{Indirect Prompt-Injection Benchmarks}
\label{sec:eval_injection}

Indirect prompt injection is the failure mode the privilege invariant is designed to prevent: a malicious instruction is planted in a data payload, such as a retrieved document or a tool response, rather than supplied by the user.
We evaluate on two published benchmarks that place the injection on different surfaces.
\textbf{OPI}~\cite{liu2024formalizing} injects into the data of classic NLP tasks and defines the five-attack taxonomy used throughout this evaluation (naive, escape, ignore, fake-completion, and combine), together with the attack-success metric.
\textbf{ASB}~\cite{zhang2025agent} moves the attack into an agent's tool observations, which is the surface that matters for tool-using deployments; we report its observation-injection arm.

We report attack success rate (ASR), the fraction of adversarial prompts in which the injected instruction is followed; lower is better.
For the LLMON models the injected payload arrives inside a ``\texttt{data}'' block reached through ``\texttt{exec<|.|>input}'', so following it would require acting on a span the model was trained to read rather than execute; the instruct baselines receive the same payloads under conventional chat formatting.


\subsubsection{Adversarial Red-Teaming}
\label{sec:eval_redteam}

The two benchmarks above measure known, published attacks.
Separately from them, and going beyond what a fixed benchmark can surface, our security team red-teamed both models with ARES~\citep{ares2025}, a robustness-evaluation harness that expands 808 seed prompts from five datasets (Purple~\citep{bhatt2023purple}, Gandalf~\citep{pfister2025gandalf}, and TensorTrust~\citep{toyer2024tensor} for direct attacks; Alpaca and a safety-policy set for indirect ones) into approximately 15k attack prompts by mutating each seed through eight attack strategies organized by attacker sophistication, from basic (asking directly) through intermediate (append, ignore, escape-separation, completion-real, append-and-prepend) to advanced.
The advanced arm is a multi-turn variant of Crescendo~\citep{russinovich2025great}, in which an attacker model generates each next turn and a judge decides whether the adversarial objective was met; we use DeepSeek-V3.2~\citep{deepseekai2025deepseekv3technicalreport} as the attacker, \textit{GPT-OSS}~\citep{openai2025gptoss120b} (120B) as the judge, and a budget of 20 turns.
Results are reported as \emph{robustness} ($1-\text{ASR}$), so that higher is better, which is the convention of the harness; note that this is the opposite direction to Table~\ref{tab:main_asr_results}.


\subsubsection{Attention-Based Interpretability Metrics}
\label{sec:attn_metrics}

Beyond output-level evaluation, we introduce three per-layer attention metrics designed to provide mechanistic insight into model behavior.
These metrics quantify how attention from the final generated token is distributed across target and distractor spans.
Let $A(S)$ denote the total attention mass assigned to a span set $S$.
We define $\mathcal{I}_t$ and $\mathcal{I}_d$ as the target and distractor instruction spans, respectively, and $\mathcal{D}_t$ and $\mathcal{D}_d$ as the corresponding target and distractor data spans.

The \textit{Instruction Focus Ratio} (IFR) measures the proportion of attention allocated to the target instruction relative to all instruction spans:

\begin{equation}
\text{IFR} = \frac{A(\mathcal{I}_t)} {A(\mathcal{I}_t)+A(\mathcal{I}_d)}
\label{eq:ifr}
\end{equation}

Similarly, the \textit{Data Attention Score} (DAS) quantifies the proportion of attention assigned to the target data relative to all data spans:

\begin{equation}
\text{DAS} = \frac{A(\mathcal{D}_t)}{A(\mathcal{D}_t)+A(\mathcal{D}_d)}
\label{eq:das}
\end{equation}

Finally, \textit{Leakage} (LEAK) is the share of all span attention that reaches distractor content, across both instruction and data spans.
Writing $\mathcal{S}$ for every labeled span in the prompt (the target and distractor instructions and data, together with the structural spans: the execution block, the delimiters, and the remaining prompt), LEAK is

\begin{equation}
\text{LEAK} = \frac{A(\mathcal{I}_d) + A(\mathcal{D}_d)}{A(\mathcal{S})}
\label{eq:leak}
\end{equation}

These metrics are computed per layer across all attention heads (averaged over attention heads within each KV group for grouped-query-attention models).
This enables layer-wise failure diagnosis: higher IFR and DAS values indicate stronger focus on task-relevant content, whereas larger LEAK values suggest increased attention allocated to irrelevant or competing information.

%% file: tex/05_results.tex
\section{Results and Discussion}
\label{sec:results}


\subsection{General Capabilities}

Table~\ref{tab:main_results} compares LLMON-structured post-training against the \textit{Instruct} chat-template baselines at both scales under matched full fine-tuning.
The objective here is parity rather than improvement: LLMON training is meant to install an execution-binding behavior, not to make the model better at mathematics or code.
Averaged over the six capability categories, the 3B model is unchanged (66.7 both ways) and the 8B model loses 1.4 points (74.2 to 72.8), with mathematics at 8B the largest single regression (84.1 to 79.6).
LLMON improves on Code and Multilingual at 3B.
Safety improves at both scales, by 8.8 points at 3B and 6.2 at 8B, which is where the structural supervision is expected to show.

\begin{table*}[t]
\centering
\small
\begin{tabular}{llccccccc}
\toprule
\textbf{Model} & \textbf{Training} & \textbf{General} & \textbf{Alignment} & \textbf{Math} & \textbf{Code} & \textbf{Tool-Calling} & \textbf{Multilingual} & \textbf{Safety} \\
\midrule
\multirow{2}{*}{Granite-4.1-3B} & Instruct & \textbf{57.91} & \textbf{82.30} & \textbf{75.19} & 64.19 & \textbf{60.80} & 59.89 & 80.89 \\
 & LLMON & 57.16 & 81.02 & 73.25 & \textbf{65.45} & 60.71 & \textbf{62.68} & \textbf{89.64} \\
\midrule
\multirow{2}{*}{Granite-4.1-8B} & Instruct & \textbf{64.95} & \textbf{87.06} & \textbf{84.09} & \textbf{72.38} & \textbf{68.27} & \textbf{68.68} & 84.19 \\
 & LLMON & 63.31 & 86.69 & 79.63 & 71.01 & 67.93 & 68.24 & \textbf{90.36} \\
\bottomrule
\end{tabular}
\caption{LLMON-structured post-training vs.\ \textit{Instruct} chat-template baselines under matched full fine-tuning. Each column is the unweighted mean over the benchmarks of that category (Section~\ref{sec:eval_standard}). Capability is broadly preserved (the six-category capability mean is unchanged at 3B and 1.4 points lower at 8B) while Safety improves at both scales. Best per model/category in \textbf{bold}.}
\label{tab:main_results}
\end{table*}


\subsection{Indirect Prompt Injection}
\label{sec:results_injection}

Table~\ref{tab:main_asr_results} reports attack success rate on the two injection benchmarks of Section~\ref{sec:eval_injection}.
On text-level injection (OPI) the attack essentially stops working: ASR falls from 100.0 to 0.1 at 3B and from 96.3 to 0.02 at 8B.
On agentic injection (ASB), where the payload arrives in a tool observation rather than in task data, ASR falls from 99.2 to 5.6 at 3B and from 52.8 to 3.1 at 8B.
These are reductions of roughly $17\times$ on agentic injection and three orders of magnitude on text-level injection, and they hold at both scales despite the 8B \textit{Instruct} baseline already being markedly more robust on ASB than the 3B one.
The mechanism is structural rather than a learned refusal: the injected instruction occupies a ``\texttt{data}'' block that no execution pointer references, so acting on it would require the model to execute a span it was trained to read.
Section~\ref{sec:results_attention} examines what this looks like inside the network.

\begin{table}[t]
\centering
\small
\begin{tabular}{llcc}
\toprule
\textbf{Model} & \textbf{Training} & \textbf{OPI $\downarrow$} & \textbf{ASB $\downarrow$} \\
\midrule
\multirow{2}{*}{Granite-4.1-3B} & Instruct & 100.0 & 99.2 \\
 & LLMON & \textbf{0.1} & \textbf{5.6} \\
\midrule
\multirow{2}{*}{Granite-4.1-8B}  & Instruct & 96.3 & 52.8 \\
 & LLMON & \textbf{0.02} & \textbf{3.1} \\
\bottomrule
\end{tabular}
\caption{LLMON-structured post-training vs.\ \textit{Instruct} chat-template baselines, measured by \textbf{attack success rate (ASR, \%)} on two published indirect prompt-injection benchmarks: OPI (text-level injection) and ASB (agentic tool-observation injection). \textbf{Lower is better}: ASR is the fraction of adversarial prompts that succeed, so a smaller value means a more robust model. Best (lowest ASR) per model/benchmark in \textbf{bold}.}
\label{tab:main_asr_results}
\end{table}


\subsection{Adversarial Red-Teaming}
\label{sec:results_redteam}

Independently of the published benchmarks of Section~\ref{sec:results_injection}, Table~\ref{tab:redteam} reports the ARES red-teaming campaign of Section~\ref{sec:eval_redteam}, which uses its own seed datasets and attack strategies and separates results by where the adversarial instruction enters.
Scores are robustness ($1-\text{ASR}$), so higher is better.

On \emph{indirect} attacks the gain is large and is what the design predicts: robustness rises from 2.8\% to 84.6\% at 3B and from 0.7\% to 79.8\% at 8B, because the payload lands in a span that no execution block designates as an operable instruction.
\emph{Direct} and \emph{multi-turn} attacks also improve, roughly doubling at both scales, but remain far weaker in absolute terms (32-37\%).
That asymmetry is expected rather than disappointing.
When the adversarial instruction \emph{is} the user's own instruction there is no instruction-data boundary left to enforce, and an attacker who steers a conversation across twenty turns is not smuggling an instruction past the binding but building towards a request the model accepts on its own terms.
The privilege invariant is therefore a defense against untrusted \emph{input}, and should be composed with defenses of a different kind rather than relied upon alone.

\begin{table}[t]
\centering
\small
\begin{tabular}{llccc}
\toprule
\textbf{Model} & \textbf{Training} & \textbf{\scriptsize{Indirect}} & \textbf{\scriptsize{Direct}} & \textbf{\scriptsize{Multi-turn}} \\
\midrule
\multirow{2}{*}{Granite-4.1-3B} & Instruct & 2.8 & 17.6 & 15.0 \\
 & LLMON & \textbf{84.6} & \textbf{37.1} & \textbf{28.0} \\
\midrule
\multirow{2}{*}{Granite-4.1-8B} & Instruct & 0.7 & 9.2 & 15.0 \\
 & LLMON & \textbf{79.8} & \textbf{32.7} & \textbf{37.0} \\
\bottomrule
\end{tabular}
\caption{Adversarial red-teaming with ARES, reported as \textbf{robustness ($1-\text{ASR}$, \%)}, \textbf{higher is better}, the opposite direction to Table~\ref{tab:main_asr_results}. Indirect attacks use StruQ-style payloads planted in input, averaged over in-distribution and out-of-distribution splits (ID: 90.3/83.6, OOD: 79.0/75.9 for 3B/8B); direct attacks place the adversarial instruction in the user turn; multi-turn is a Crescendo variant with a 20-turn budget. Best per model/column in \textbf{bold}.}
\label{tab:redteam}
\end{table}


\subsection{Attention Interpretability Analysis}
\label{sec:results_attention}

Both the injection reductions and the capability picture are consistent across the two scales tested, so the remainder of this section drills into \emph{how} that robustness arises.
Using the per-layer attention metrics defined in Section~\ref{sec:attn_metrics}, we analyze the \textit{Granite-4.1-3B} pair (the LLMON-trained model and its \textit{Instruct} counterpart) on distractor prompts in which several instructions, and in most cases several data segments, coexist while the ``\texttt{exec}'' block binds only one pair.
We first report the per-layer behavior on a representative example to expose the mechanism, then aggregate over all examples to verify that the trends are not driven by ordering.

\paragraph{Where the binding lives.}
Both models are measured at the same layers, and the separation is largest in the late-middle of the stack rather than at the output.
Table~\ref{tab:attn_aggregate} reports the mean of each metric over the distractor examples at five representative layers.
The LLMON-trained model holds a mean IFR of 86-87\% at layers~25 and~28 against 26-38\% for the \textit{Instruct} model, with mean leakage of 6-9\% against 35-45\%.
At the final layer both models collapse towards chance (IFR 28.0\% vs 17.2\%), which is why we report the span distribution at layer~31 rather than at layer~39: the execution binding is a mid-to-late-stack computation, and reading it off the last layer alone would understate it for both models.

\begin{table}[t]
\centering
\small
\begin{tabular}{lcccccc}
\toprule
 & \multicolumn{2}{c}{\textbf{IFR} $\uparrow$} & \multicolumn{2}{c}{\textbf{DAS} $\uparrow$} & \multicolumn{2}{c}{\textbf{LEAK} $\downarrow$} \\
\cmidrule(lr){2-3}\cmidrule(lr){4-5}\cmidrule(lr){6-7}
\scriptsize{Layer} & \scriptsize{LLMON} & \scriptsize{Inst.} & \scriptsize{LLMON} & \scriptsize{Inst.} & \scriptsize{LLMON} & \scriptsize{Inst.} \\
\midrule
25 & \textbf{86.5} & 25.8 & \textbf{59.1} & 38.0 & \textbf{8.6} & 44.9 \\
28 & \textbf{87.2} & 37.7 & \textbf{62.2} & 36.6 & \textbf{6.3} & 34.9 \\
31 & \textbf{63.8} & 15.3 & \textbf{53.9} & 37.2 & \textbf{12.2} & 40.6 \\
34 & \textbf{75.8} & 26.3 & \textbf{57.9} & 35.6 & \textbf{6.4} & 47.2 \\
39 & \textbf{28.0} & 17.2 & \textbf{47.0} & 40.2 & \textbf{13.7} & 39.7 \\
\bottomrule
\end{tabular}
\caption{Attention metrics (\%) averaged over the distractor examples, at five representative layers, for \textit{Granite-4.1-3B} LLMON-trained vs.\ \textit{Instruct}. The LLMON model holds higher instruction focus and lower leakage at every layer measured; the gap is widest at layers~25-34 and narrows at the output.}
\label{tab:attn_aggregate}
\end{table}

\begin{figure}[t]
\centering
\includegraphics[width=\columnwidth]{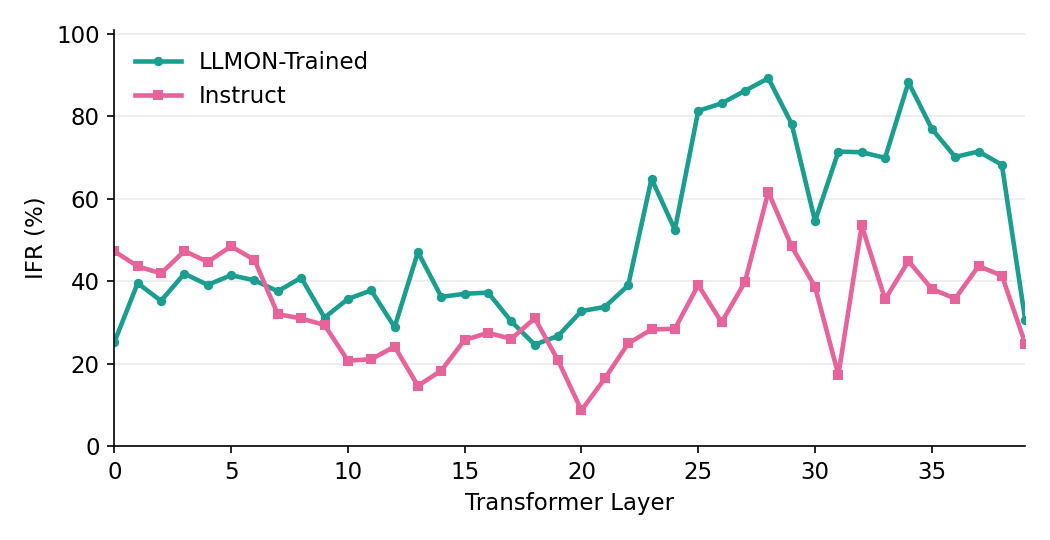}
\caption{Instruction Focus Ratio (IFR) per transformer layer for the LLMON-trained \textit{Granite-4.1-3B} (teal) vs.\ its \textit{Instruct} counterpart (pink) on a distractor example. The two are indistinguishable through layer~20 and separate sharply from layer~22, with the LLMON model sustaining 55-89\% across layers~25-38. Both fall at the final layer.}

\label{fig:ifr_per_layer}
\end{figure}

\paragraph{Layer-wise Instruction Focus Ratio.}
Figure~\ref{fig:ifr_per_layer} shows the IFR computed at each transformer layer for both models on a representative distractor example, in which three instructions (``summarize'', ``count the words'', ``translate'') and two data segments coexist and the ``\texttt{exec}'' block binds only the word-counting instruction.
The two models are indistinguishable through the first twenty layers, both fluctuating around 25-45\%.
They separate sharply from layer~22: the LLMON-trained model rises into a 55-89\% band across layers~25-38, peaking at 89\% at layer~28, whereas the \textit{Instruct} model peaks at 62\% and spends most of that range below 45\%.
The implication is that the binding is not a surface property of the prompt but a computation the network performs, and that LLMON training installs it in a specific region of the stack rather than shifting attention uniformly.

\begin{figure}[t]
\centering
\includegraphics[width=\columnwidth]{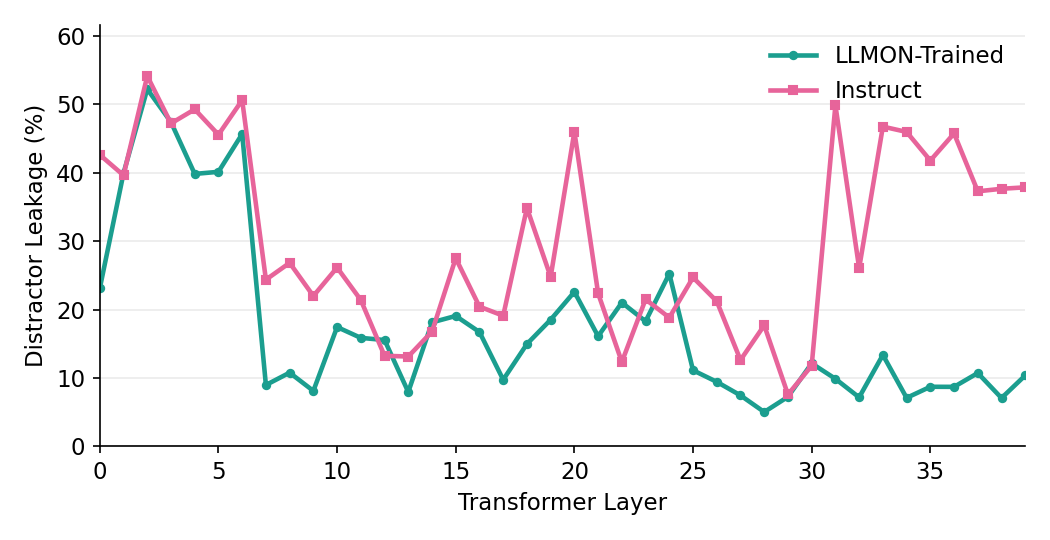}
\caption{Distractor leakage (LEAK) per transformer layer. Both models leak heavily in the first six layers; from layer~7 the LLMON-trained model (teal) settles into a 5-25\% band while the \textit{Instruct} model (pink) remains between 20\% and 50\% and rises again in the late layers.}
\label{fig:leak_per_layer}
\end{figure}

\paragraph{Layer-wise Distractor Leakage.}
Figure~\ref{fig:leak_per_layer} presents the complementary LEAK metric across layers.
Both models leak heavily in the first six layers, peaking above 50\%, which is consistent with early layers aggregating context indiscriminately.
From layer~7 the LLMON-trained model drops to a 5-25\% band and settles below 15\% after layer~25, while the \textit{Instruct} model stays in the 20-50\% range throughout and rises again in the last ten layers.
Distractor suppression is therefore not a property of the input encoding (where the two models behave alike) but something acquired in the same region of the stack where instruction focus is established.

\begin{figure*}[t]
\centering
\includegraphics[width=0.85\textwidth]{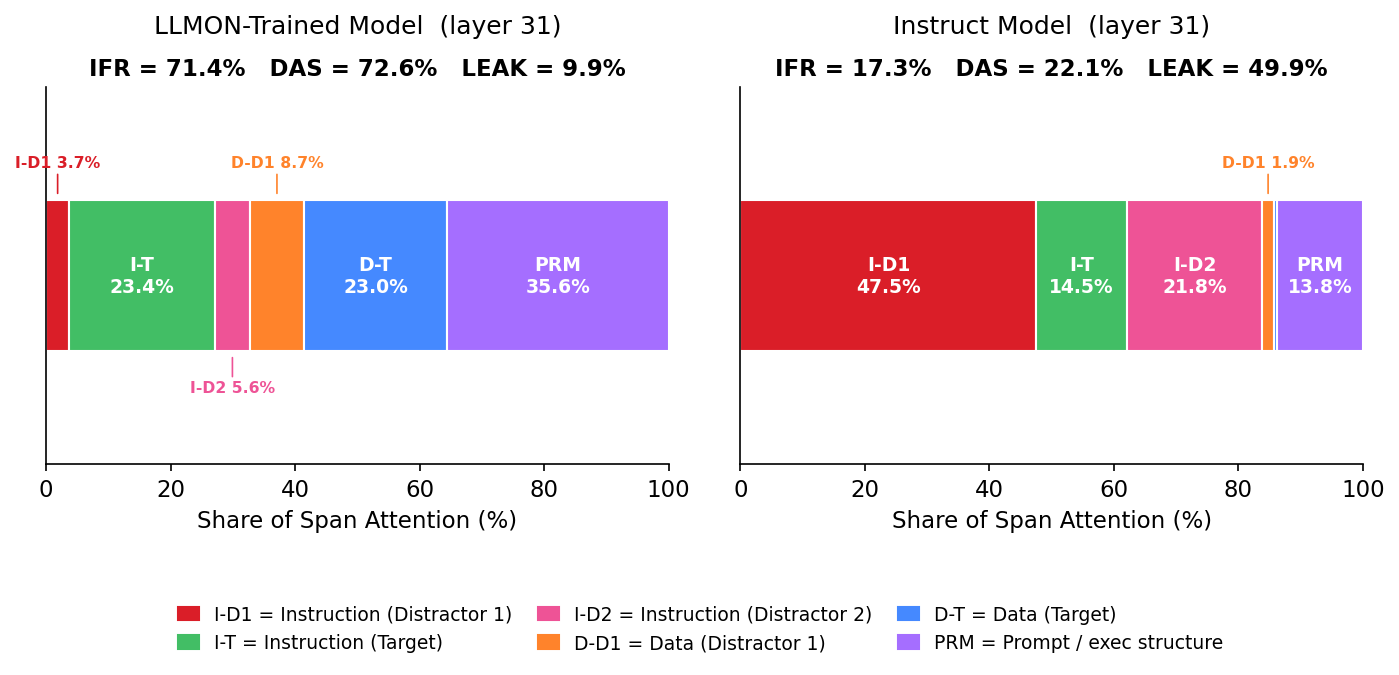}
\caption{Attention metrics and per-span attention distribution at layer~31 for the LLMON-trained \textit{Granite-4.1-3B} (left) vs.\ its \textit{Instruct} counterpart (right). The LLMON model divides its attention between the bound target instruction (I-T) and target data (D-T); the \textit{Instruct} model's largest single recipient is a distractor instruction (I-D1, 47.5\%).}
\label{fig:span_dist}
\end{figure*}

\paragraph{Attention span distribution.}
Figure~\ref{fig:span_dist} presents the per-span attention distribution and the three metrics at layer~31 for both models.
The LLMON-trained model splits its span attention between the target instruction (I-T, 23.4\%) and the target data it is bound to (D-T, 23.0\%), with 35.6\% on the prompt and execution structure and only 5.6\% on the largest distractor instruction: IFR = 71.4\%, DAS = 72.6\%, LEAK = 9.9\%.
The \textit{Instruct} model inverts this. Its single largest recipient is a distractor instruction (I-D1, 47.5\%), with a second distractor taking 21.8\% and the target instruction only 14.5\%; target data attention is negligible.
Its metrics record the failure directly: IFR = 17.3\%, DAS = 22.1\%, LEAK = 49.9\%.
The two models receive identical token sequences, so the difference is entirely in what the network does with them.

\begin{figure}[t]
\centering
\includegraphics[width=\columnwidth]{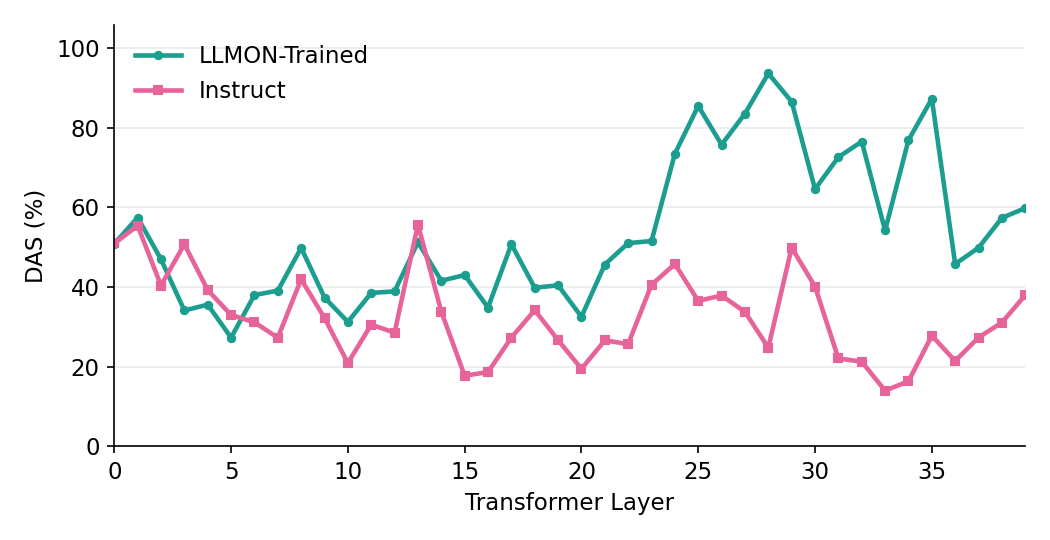}
\caption{Data Attention Score (DAS) per transformer layer for the LLMON-trained \textit{Granite-4.1-3B} (teal) vs.\ its \textit{Instruct} counterpart (pink). DAS separates in the same region of the stack as IFR and LEAK, rising into a 46-94\% band across layers~23-38 for the LLMON model.}
\label{fig:das_per_layer}
\end{figure}

\paragraph{Layer-wise Data Attention Score.}
Figure~\ref{fig:das_per_layer} completes the picture with DAS.
It follows the same shape as IFR: the two models track each other through the first twenty layers, then the LLMON-trained model rises into a 46-94\% band across layers~23-38 while the \textit{Instruct} model stays between 14\% and 56\%.
All three metrics therefore separate in the same region of the stack, which is what one expects if a single mechanism of resolving the execution binding is responsible for all three.

\begin{figure*}[t]
\centering
\includegraphics[width=0.85\textwidth]{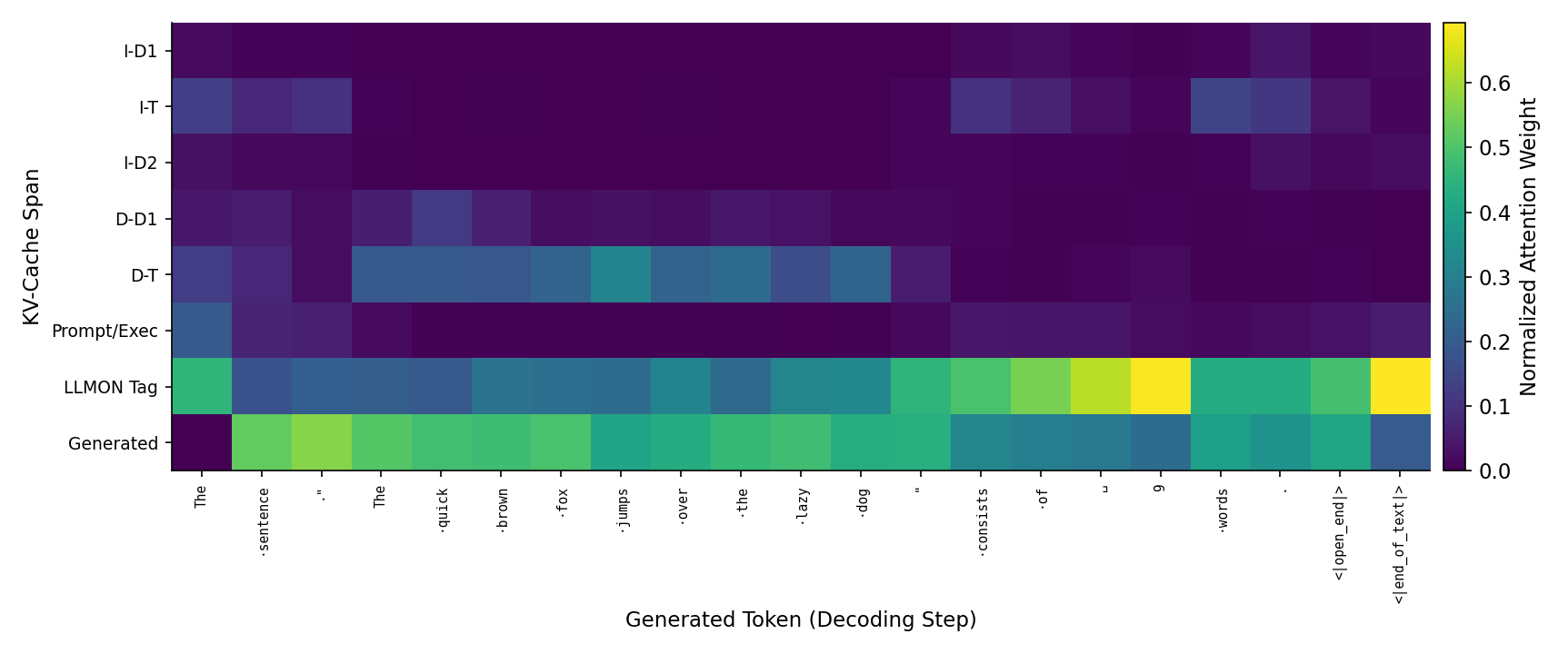}
\caption{Attention correlation weights between each generated token (columns, in decoding order) and the KV-cache spans (rows) for the LLMON-trained \textit{Granite-4.1-3B} at layer~31. Brighter cells indicate stronger attention. The target data span (D-T) activates while the model quotes the sentence it was asked to count, and the distractor spans (I-D1, I-D2, D-D1) remain near zero throughout decoding.}
\label{fig:span_attention_flow}
\end{figure*}

\paragraph{Token-level attention flow.}
Figure~\ref{fig:span_attention_flow} traces how attention is allocated during generation, showing the correlation between each generated token and the KV-cache spans for the LLMON-trained model at layer~31.
The target data span (D-T) activates precisely while the model is quoting the sentence it was asked to count, and falls away once it moves on to report the count; the target instruction span (I-T) carries weight at the opening tokens and again as the answer is formed.
The distractor spans (I-D1, I-D2, D-D1) stay near zero throughout decoding.
This token-resolved view complements the per-layer metrics: the model does not merely arrive at the right binding by the end of the forward pass, it holds it step by step as it produces the answer.

%% file: tex/06_conclusion.tex
\section{Conclusion}
\label{sec:conclusion}


We presented a structured post-training methodology that uses LLMON's explicit instruction-data separation and identifier-based execution binding to improve controlled execution in LLMs.
Our evaluation on two \textit{Granite-4.1} base models demonstrates that:
(i) LLMON-structured supervision reduces attack success rates on indirect prompt-injection benchmarks from 53-100\% to 0.02-5.6\%, without catastrophic degradation on standard capability measures;
(ii) the effect holds at both 3B and 8B scale, and extends to adversarial red-teaming on indirect attacks, though direct and multi-turn attacks remain substantially harder;
and (iii) attention-based interpretability metrics (IFR, DAS, LEAK) provide actionable per-layer failure diagnosis, revealing how structural training reshapes internal attention patterns.
These findings suggest that explicit execution structure can be reliably taught to LLMs through standard post-training, offering a training-time complement to inference-time safety mechanisms.

%% file: tex/07_limitations.tex
\section*{Limitations}


We discuss several important limitations of this work.
First, our evaluation covers two model scales (3B and 8B) within a single model family and generation.
While LLMON's structural supervision is architecture-agnostic, it remains to be validated whether the same magnitude of improvement transfers to other pre-training corpora and architectures, and to larger scales where models may already exhibit stronger implicit instruction-following.
Second, LLMON's delimiter tokens increase context length by approximately 7-15\% depending on the example complexity.
We do not systematically quantify the impact on inference latency and throughput, though sequence packing at training time largely amortizes this cost.
Third, the privilege invariant addresses \emph{indirect} injection, where the adversarial instruction arrives as input.
It offers far less protection when the adversarial instruction \emph{is} the user's instruction, since there is then no instruction-data boundary to enforce: red-teaming robustness reaches only 33-37\% on direct attacks and 28-37\% on multi-turn attacks, against 80-85\% on indirect ones, and it is a learned behavior rather than an enforced guarantee: it substantially reduces, but does not eliminate, the chance that an instruction embedded in data is followed, and should be one layer of a defense rather than the only one.
Fourth, while our capability evaluation includes multilingual benchmarks, the LLMON training corpus itself is predominantly English, and we have not characterized how tag overhead behaves on morphologically rich languages or on tasks with very long documents.
Finally, our IFR/DAS/LEAK metrics provide correlation between attention patterns and output quality, but we do not establish a causal mechanism.
Attention weights are an imperfect proxy for information flow in transformers, and the relationship between attention-level metrics and generation-level performance may not be monotonic.

%% file: tex/08_appendix.tex
\section{LLMON Examples}
\label{sec:appndx_examples}

\subsection{Multi-instruction and Multi-data Distractors}
\label{sec:appndx_distractor}

Figure~\ref{lst:llmon_distractor_example} illustrates a distractor example: three instructions and two data segments coexist, but the ``\texttt{exec}'' block unambiguously binds only one instruction-data pair.
The model must learn to attend selectively to the referenced segments and ignore all others.

\begin{figure*}[t]
\centering
\begin{lrbox}{\llmonbox}
\begin{minipage}{0.97\linewidth}
\begin{lstlisting}[
    escapechar=@,
    basicstyle=\ttfamily\tiny,
    numbers=none,
    breaklines=true,
    breakindent=0pt,
    showstringspaces=false
]
@\colorbox{pink}{<|open|>}@@\colorbox{lime}{instr<|:|>êí}@@\colorbox{pink}{<|close|>}@
Summarize the text in one short sentence
@\colorbox{pink}{<|open\_end|>}@
@\colorbox{pink}{<|open|>}@@\colorbox{lightyellow}{data<|:|>¡¿}@@\colorbox{pink}{<|close|>}@
The smart rabbit escaped the field
@\colorbox{pink}{<|open\_end|>}@
@\colorbox{pink}{<|open|>}@@\colorbox{yellow}{data<|:|>Äî}@@\colorbox{pink}{<|close|>}@
The quick brown fox jumps over the lazy dog
@\colorbox{pink}{<|open\_end|>}@
@\colorbox{pink}{<|open|>}@@\colorbox{green}{instr<|:|>íä}@@\colorbox{pink}{<|close|>}@
Count the number of words in the given sentence
@\colorbox{pink}{<|open\_end|>}@
@\colorbox{pink}{<|open|>}@@\colorbox{lime}{instr<|:|>Úê}@@\colorbox{pink}{<|close|>}@
Translate the text into French
@\colorbox{pink}{<|open\_end|>}@

@\colorbox{pink}{<|open|>}@@\colorbox{orange}{exec<|:|>áÂ}@@\colorbox{pink}{<|close|>}@
  @\colorbox{pink}{<|open|>}@@\colorbox{orange}{exec<|.|>instruction}@@\colorbox{pink}{<|close|>}@@\colorbox{green}{instr<|:|>íä}@@\colorbox{pink}{<|open\_end|>}@
  @\colorbox{pink}{<|open|>}@@\colorbox{orange}{exec<|.|>input}@@\colorbox{pink}{<|close|>}@@\colorbox{yellow}{data<|:|>Äî}@@\colorbox{pink}{<|open\_end|>}@
  @\colorbox{pink}{<|open|>}@@\colorbox{orange}{exec<|.|>output}@@\colorbox{pink}{<|close|>}@@\colorbox{lightblue}{assistant<|:|>ðã}@@\colorbox{pink}{<|open\_end|>}@
@\colorbox{pink}{<|open\_end|>}@

@\colorbox{pink}{<|open|>}@@\colorbox{lightblue}{assistant<|:|>ðã}@@\colorbox{pink}{<|close|>}@
The sentence "The quick brown fox jumps over the lazy dog" consists of 9 words.
@\colorbox{pink}{<|open\_end|>}@@\colorbox{lightgray}{<|end\_of\_text|>}@
\end{lstlisting}
\end{minipage}
\end{lrbox}
\fbox{\usebox{\llmonbox}}
\caption{Multi-instruction distractor example. Three instructions and two data segments coexist; the bound target instruction and data are shown in green and yellow, the distractors in lime and light-yellow. The exec block (orange) references only ``\texttt{instr<|:|>íä}'' with input ``\texttt{data<|:|>Äî}''. The model must execute only the word-counting instruction, ignoring the summarization and translation distractors. Other special tokens are highlighted in light-pink.}
\label{lst:llmon_distractor_example}
\end{figure*}